# Segmentation of optic disc, fovea and retinal vasculature using a single convolutional neural network

Jen Hong Tan[1], U. Rajendra Acharya [1,2,3], Sulatha V. Bhandary[4], Kuang Chua Chua[1], Sobha Sivaprasad[5]

[1]Department of Electronics and Computer Engineering, Ngee Ann Polytechnic, Singapore.
[2]Department of Biomedical Engineering, School of Science and Technology, SIM University, Singapore.
[3]Department of Biomedical Engineering, Faculty of Engineering, University of Malaya, Malaysia
[4]Department of Ophthalmology, Kasturba Medical College, Manipal India 576104
[5]NIHR Moorfields Biomedical Research Centre, London, UK

*Abstract*

We have developed and trained a convolutional neural network to **automatically and simultaneously** segment optic disc, fovea and blood vessels. Fundus images were normalized before segmentation was performed to enforce consistency in background lighting and contrast. For every effective point in the fundus image, our algorithm extracted three channels of input from the point's neighbourhood and forwarded the response across the 7-layer network. The output layer consists of four neurons, representing background, optic disc, fovea and blood vessels. In average, our segmentation correctly classified 92.68% of the ground truths (on the testing set from Drive database). The highest accuracy achieved on a single image was 94.54%, the lowest 88.85%. A single convolutional neural network can be used not just to segment blood vessels, but also optic disc and fovea with good accuracy.

## 1. Introduction

Automated segmentation of retinal vasculature and optic disc are well-studied in literature, but less so for fovea. Blood vessels and optic disc on fundus images generally have distinct boundaries (see **Figure 1**), and therefore exhibit a clearly delineated region. Fovea, however, shows no border.

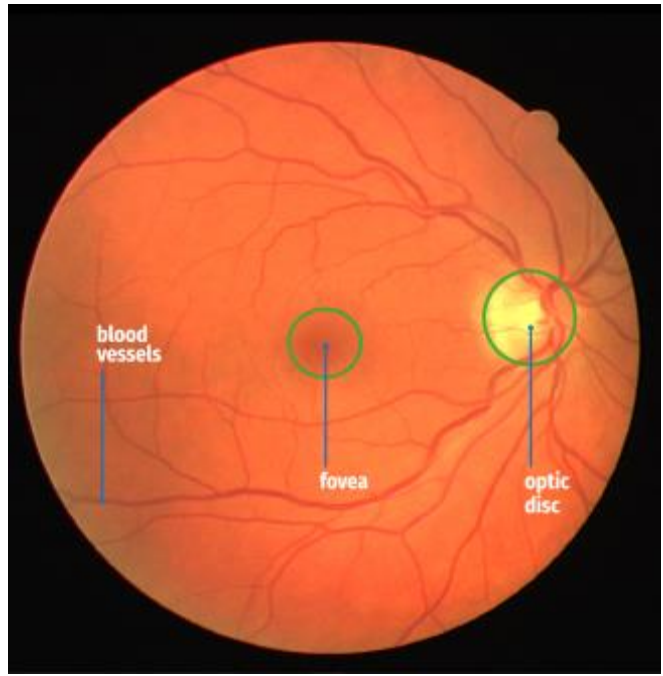

**Figure 1.** The blood vessels, optic disc and fovea in a fundus image.

To date there is no single study that works to ***automatically*** and ***simultaneously segment*** optic disc, fovea and vasculature. Most of the proposed automated segmentation focused only on either one of them, with majority of the works dedicated to vasculature.

Fraz et al. [1] has made an extensive review on the works that solely segment vasculature, in which they have divided the proposed solutions in literature into seven categories: pattern recognition [2-4], matched filtering [5,6], vessel tracking/tracing [7], mathematical morphology [8,9], multiscale approach [10,11], model based approach [12,13] and parallel/hardware approach [14]. Most of the methodologies reported their performances in terms of accuracy, sensitivity, specificity and area under the curve, and majority of them were evaluated on DRIVE database [1].

On the other hand, for optic disc most of the proposed solutions [15] used methods either based on intensity [16,17], template matching [18,19] or vasculature structure [20,21]. These published works can also be categorized into two types [21]: methods that locate the optic disc but do not segment optic disc [16,22], and methods that segment optic disc [17,18]. When segmentation was actually performed, overlap measure was reported for performance.

However, on fovea segmentation, overlap measure was never reported, for nearly all of the published works on fovea segmentation were designed to locate the position of the fovea, not to segment the region of interest. When the fovea is located, the positions acquired can be used for the assessment of diabetic retinopathy, as detailed in the work by Medhi and Dandapat [23].

The solutions proposed in literature [24] often require the information of optic disc and/or blood vessels, and in some methodologies, the fovea was located with the knowledge of the specific branches of vasculature and the position of the optic disc. For example, Li and Chutatape [25] have used a modified active shape model to get the contours converged on the outer rim of optic disc and its neighbouring vascular arch. From there the algorithm determined the position of the fovea.

And it is not only Li and Chutatape [25] that used shape model to look for the optic disc and fovea. In the work by Niemeijer et al. [20], they also used a point distribution model to locate similar objects. The algorithm proposed did not delineate any boundary, and it was only three years later, in another paper [26], a kNN-regressor and a circular template were proposed by the authors to enclose the boundary of optic disc and fovea. Even then they only evaluated their algorithm against the position of the optic disc and fovea, not the amount of area overlapped with a ground truth.

On the other hand, Sinthanayothin et al. [16] has in a study developed methods to segment blood vessels and *locate* the position of optic disc and fovea automatically. The blood vessels were identified using a three-layer perceptron. The optic disc was localized based on the intensity variation among adjacent pixels, and then the location of fovea was determined at a distance 2.5 times the diameter of the optic disc from the position of the optic disc.

Besides, Narasimha-Iyer et al. [27] have proposed a solution to segment vasculature, optic disc and fovea using three separate algorithms. However, as the focus of the paper was to detect longitudinal changes in fundus image, not segmentation, they only visually evaluated the segmentation of optic disk and fovea, which they found acceptable [27].

None of the above studies performed a *simultaneous segmentation* of vasculature, optic disc and fovea. When vasculature, optic disc and fovea were actually segmented, separate algorithms were used. Furthermore, these algorithms work by assuming vasculature, optic disc and fovea are present in the image of interest. They cannot indicate the absence of a feature when the feature is not available in the image.

In this paper we are going to propose a solution not just to locate, but to *simultaneously segment* vasculature, optic disc *and* fovea. We use a 7-layer convolutional neural network (CNN) to classify every pixel into one of the four classes: background, blood vessels, optic disc and fovea. When any of the features is not available in the image, our algorithm is unlikely to produce any segmentation of the missing feature.

CNN has been adopted recently for segmentation on fundus images [3,28,29]. Among the works, Wang et al. [3] has produced the best accuracy on DRIVE database [30]. The team has used a 7-layer CNN together with random forest algorithm to extract retinal vasculature. However, in their work CNN was used only as a trainable feature extractor; the classification was in fact performed by an ensemble of random forest.

The input to their neural network is a 25 x 25 image patch, extracted from green channel. But in our case, to segment fovea and optic disc, a larger input, or an input of other form is definitely needed to discriminate a pixel among the four classes.

The reasons can be seen in **Figure 2**. On the first row of the figure, there are two frames (each of them sizes 25 x 25) that should be classified as fovea and another two as background. But there are almost no discernible differences among the four (except the colour, which is not a useful clue). On the other hand, although optic disc is generally considered to have a distinct appearance that separates itself from the rest of the image, this is not always the case when we frame part of the region under a box of 25 x 25, as illustrated in the second row of **Figure 2**.

However, an input of size say 101 x 101 is not really a good choice, for the training will take too much time and too many computer memories. To make our segmentation possible, we propose an input of size 33 x 33, but with three channels. For a pixel at location $(x, y)$, the first channel of the input is a 7 x 7 neighbourhood around the pixel but scaled to a size of 33 x 33. The second channel is just a 33 x 33 neighbourhood surrounding the pixel. The final channel is a region spanning 165 x 165, with the pixel as its center but scaled down to a size of 33 x 33. Further details on the architecture will be elaborated in section 3.

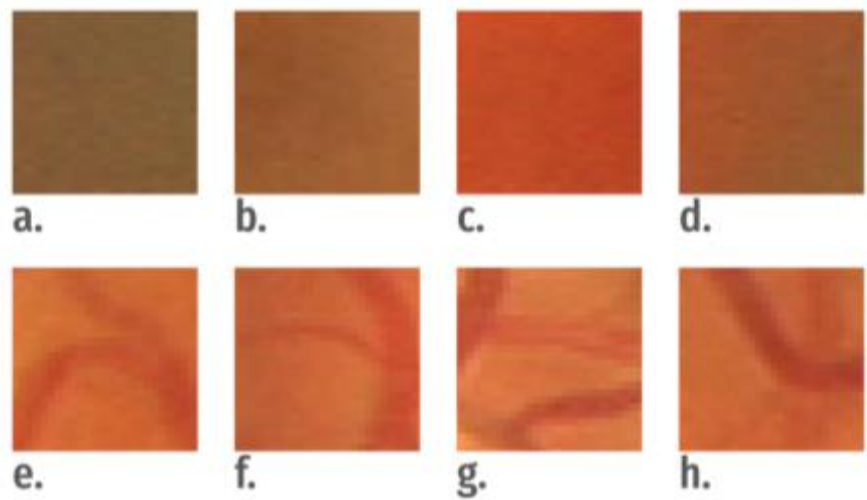

**Figure 2.** Pixels at fovea region can be indistinguishable from background pixels if we only look at a frame of 25 x 25 centered around the pixel of interest. In the first row, frame b, c are extracted from fovea region; a,

d from background. Similar confusion can also arise for pixels from optic disc and regions other than optic disc. In the second row, frame f, g are taken from optic disc region; e, h from non-optic-disc region.

## 2. Materials

We developed and tested our algorithm on the Drive database [30], which is available for public. The database provides a total of 40 fundus images, in which 33 of them exhibits no symptom of diabetic retinopathy, with the rest shows traits of mild early diabetic retinopathy. The database also provides every image a mask, so that non-fundus regions can be excluded during analysis.

Canon CR 5 non-mydriatic 3CCD camera was used to take the images. The field of view is 45 degree. Each image sizes 565 x 584, and the entire database is divided into a training and a testing set, each consists of 20 images.

We trained our CNN on the training set and tested it on the testing set. The ground truths on the vasculature for the training and the testing set were provided by the database. However, we used only the vasculature ground truths provided in the folder '1st_manual' for testing. For optic disc and fovea, all the ground truths were delineated by an ophthalmologist who has a clinical experience of more than 10 years.

**3. Methods**

3.1 Normalization

We normalized the colour image before we performed any classification (see **Figure 3**). We used the procedures laid out in the Section 3.A (official manuscript)/Section 3.1 (uploaded manuscript) from [31] to do the normalization, with only a change. In the original article, the processes were done on the green channel of a fundus image to produce a ***normalized grey-scale image*** for vasculature extraction. In this study, we converted a fundus image from RGB colour space to LUV colour space, ran the procedures on the L channel (luminance channel), and with the adjusted L channel, we converted the image back to RGB colour space. In this case the product is a ***normalized colour image***, which has a wider usage than the output by the previous study.

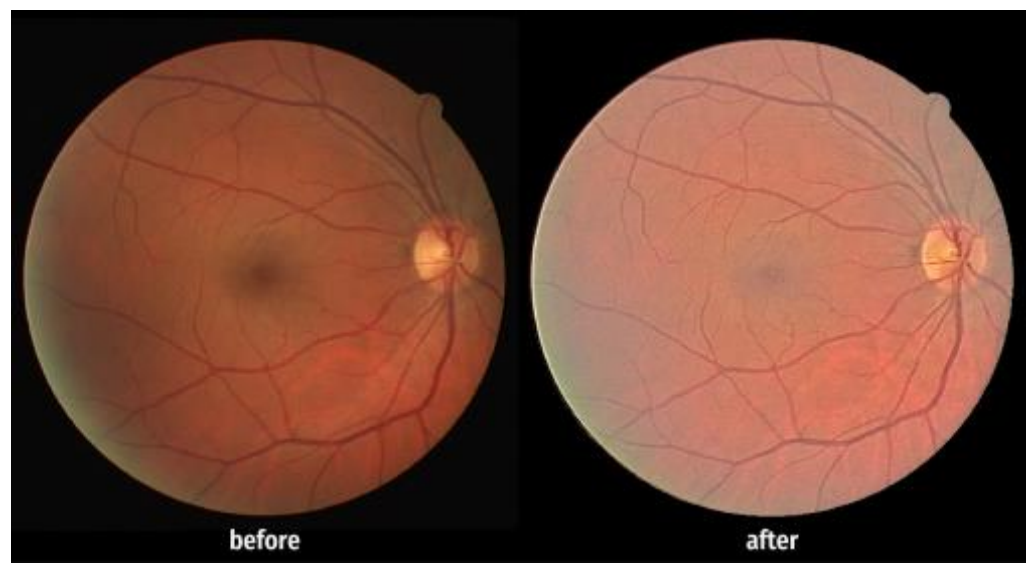

**Figure 3.** Before and after the normalization.

3.2 The architecture

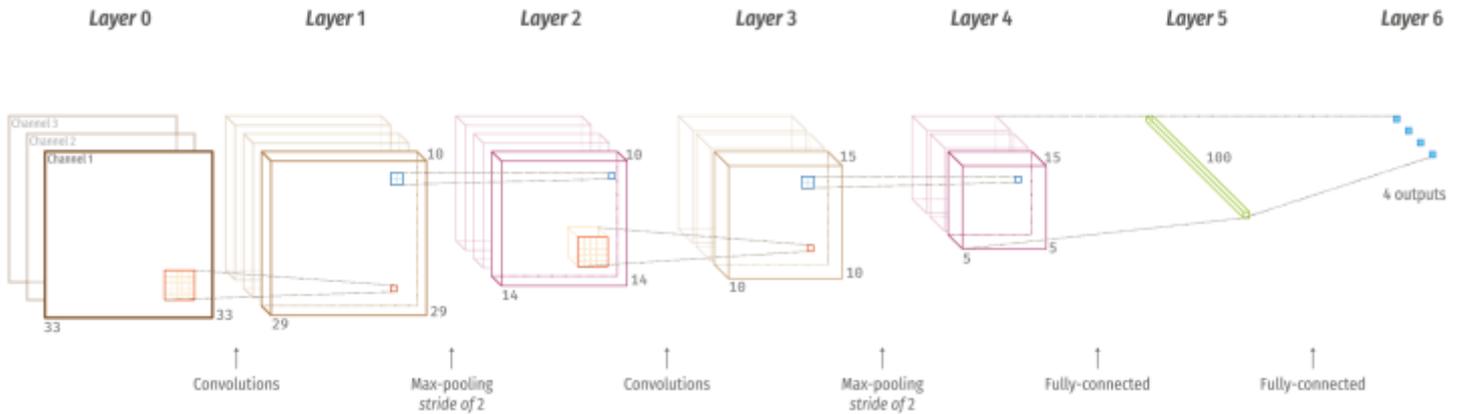

**Figure 4.** The architecture of the proposed CNN.

The structure of the proposed neural network is illustrated in **Figure 4** and detailed in **Table 1**. The input (layer 0) consists of 3 channel, each sizes 33 x 33. ***For each channel***, its 33 x 33 matrix is convolved with 10 5 x 5 filters to produce 10 feature maps (layer 1). On each feature map a max-pooling sizing 2 x 2 is applied (layer 2).

After that, the 10 feature maps are convolved with a 5 x 5 x 10 filters to a produce a 10 x 10 feature map (layer 3). And there are 15 such feature maps produced in this layer. Max-pooling of size 2 x 2 is again applied on every feature map (layer 4). Lastly, the neurons of every feature maps in layer4 (which includes output from other channels) are connected fully to a 100 neurons in layer 5, which are also fully connected to the four outputs (neurons) in layer 6.

**Table 1.** The details on the CNN structure.

| Layers | Type | No. of neurons (output layer) | Kernel size for each output feature map | stride | No. of trainable parameters |
|--------|------|-------------------------------|------------------------------------------|--------|------------------------------|
| 0 -1 | Convolution | 29 x 29 x 30 | 5 x 5 | 1 | 780 |
| 1 - 2 | Max-pooling | 14 x 14 x 30 | 2 x 2 | 2 | - |
| 2 - 3 | Convolution | 10 x 10 x 45 | 5 x 5 x 10 | 1 | 11,545 |
| 3 - 4 | Max-pooling | 5 x 5 x 45 | 2 x 2 | 2 | - |
| 4 - 5 | Fully-connected | 100 | - | - | 112,600 |
| 5 - 6 | Fully-connected | 4 | - | - | 404 |
| Total | | | | | 125,329 |

We use only green colour channel from the normalized image. It is standardized using equation 11 in [31] before we extract any patches from it. For a given pixel at $(x_i, y_i)$, the first channel to the input is made up by a 7 x 7 neighbourhood scaled up to a size of 33 x 33. For the second channel, it is formed by a neighbourhood sizing 33 x 33 around the pixel. For the final channel, a block spanning 165 x 165 centered on the pixel are extracted from the standardized green channel of the ***original*** fundus image, but scaled down to size of 33 x 33. All of the scalings are performed using bi-cubic interpolation. **Figure 5** shows the typical inputs to our CNN.

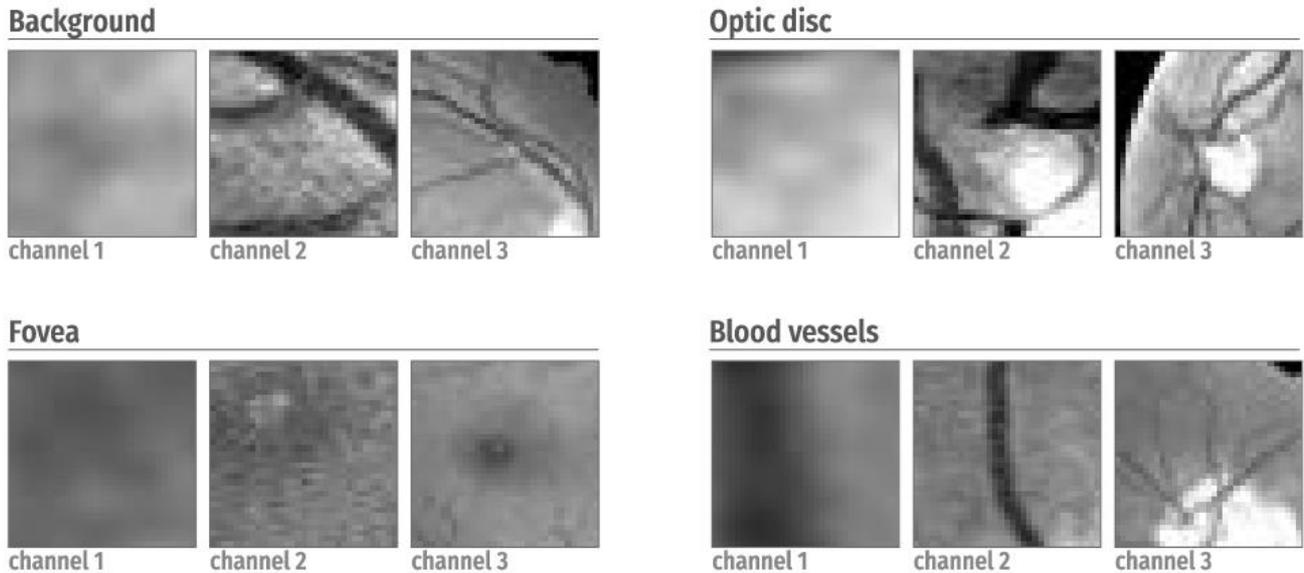

**Figure 5.** Typical inputs (layer 0) to CNN (each input consists of three channels).

Leaky rectifier linear unit (LReLU) [32] is used as the activation function for layer 1, 3, and 5. For layer 6, we use softmax function instead. Xavier initialization [33] is used for the weights of layer 1, 3 and 5. For biases, they are set to a value of 1 on layer 1, 3 and 5. On the other hand, the value of the biases at the layer 4 is randomly generated on a Gaussian distribution.

Before we settled the input size to our network on 33 x 33, we have tested several other sizes, from 9 x 9 to 45 x 45. We found that when input size of 9 x 9 was used, the computation time was much shorter, but the accuracy rarely exceeded 90%. When we increased the input size to 45 x 45, the accuracy at best hovered around 91%. And after numerous observations, we concluded that 33 x 33 gave the best accuracy.

We have also tried an input size of 71 x 71, single channel, with a network consisting of 9 layers. The computation time was much slower, which took nearly two days to complete a single epoch. The segmentation accuracy was lower than the proposed setup, and therefore the input size was not adopted.

## 3.3 Training

Standard backpropagation [34] was used to learn and stochastic gradient descent [35], with a batch size of 10, was set up to train our CNN. The weights are updated by

$$\mathbf{w}^l = \left(1 - \frac{\eta\lambda}{\varphi}\right)\mathbf{w}^{l-1} - \frac{\eta}{\kappa}\frac{\partial c}{\partial \mathbf{w}}$$

( 1 )

where $l$ is the layer number, $\eta$ is the learning rate, $\varphi$ is the total number of training samples, $\lambda$ is the regularization parameter, $\kappa$ is the batch size. $c$ denotes cost function, which in our case is log-likelihood function. Biases are updated through

$$\mathbf{b}^l = \mathbf{b}^{l-1} - \frac{\eta}{\kappa}\frac{\partial c}{\partial \mathbf{b}}$$

( 2 )

The learning rate and the regularization parameter are set to 0.01 and 0.1 respectively.

A point is said effective if it is part of the fundus in fundus image. In total there are 4,541,006 effective points in the training set (20 fundus images), of which 3,817,049 of them are categorized as background, and an amount of 569,415, 79,321 and 75,221 points are determined as blood vessels, optic disc and fovea respectively. We did not use all of the points for training, as it would take too much time and too many computer memories to run. Instead, we took only a portion of all the available points to do the training, as listed in **Table 2**.

**Table 2.** Training samples.

|  | Original amount | Selected for training | |
| --- | --- | --- | --- |
| Background | 3,817,049 | 300,000 | (randomly selected) |
| Blood vessels | 569,415 | 150,000 | (randomly selected) |
| Optic disc | 79,321 | 150,000 | (79,321+randomly select 70,679) |
| Fovea | 75,221 | 150,000 | (75,221+randomly select 74,779) |
| Total | 4,541,006 | 750,000 | |

## 3.4 Testing

At the end of every training epoch, our algorithm performs a testing on our neural network model. The testing set provided by DRIVE database consists of 4,538,439 effective points (**Table 3** shows the break-down on each class). In each class, for every four effective points, our algorithm takes one to test the model's accuracy. 40 epochs of training and testing were run. The model that gives the best performance among the 40 epochs is then picked to run a complete classification on all the available effective points from the testing set.

**Table 3.** Testing samples.

|  | No. of points | Percentage |
|---|---|---|
| Background | 3,814,982 | 84.06% |
| Blood vessels | 577,945 | 12.73% |
| Optic disc | 78,535 | 1.73% |
| Fovea | 66,977 | 1.48% |
| Total | 4,538,439 | 100.00% |

**4. Results**

We developed our algorithm/code in MATLAB without using any other library or toolbox (except MATLAB's image processing toolbox for reading image and visualization). Some of the heavy-duty processes were written in C MEX file to speed up calculation. We train our net on a workstation which has two Intel Xeon 2.20 GHz (E5-2650 v4) processor and a 512GB RAM. It typically took about **38,749** seconds to complete an epoch of training.

**Table 4.** Segmentation results on every testing image.

| Image | Total amount of effective points | Correctly classified | Percentage |
|-------|----------------------------------|----------------------|------------|
| 1 | 224,405 | 206,282 | 91.92 |
| 2 | 225,154 | 208,356 | 92.54 |
| 3 | 225,734 | 200,566 | 88.85 |
| 4 | 227,588 | 214,447 | 94.23 |
| 5 | 227,707 | 212,304 | 93.24 |
| 6 | 227,510 | 209,409 | 92.04 |
| 7 | 227,683 | 209,462 | 92.00 |
| 8 | 225,326 | 207,077 | 91.90 |
| 9 | 227,661 | 212,879 | 93.51 |
| 10 | 227,336 | 213,262 | 93.81 |
| 11 | 227,820 | 210,371 | 92.34 |
| 12 | 227,605 | 211,047 | 92.73 |
| 13 | 227,507 | 211,841 | 93.11 |
| 14 | 225,900 | 203,796 | 90.22 |
| 15 | 227,396 | 212,534 | 93.46 |
| 16 | 227,694 | 213,130 | 93.60 |
| 17 | 225,885 | 207,139 | 91.70 |
| 18 | 227,612 | 212,840 | 93.51 |
| 19 | 227,392 | 214,987 | 94.54 |
| 20 | 227,524 | 214,534 | 94.29 |
| | 4,538,439 | 4,206,263 | (mean) 92.68 |

We performed the complete testing on a Mac Pro, which has two Intel 2.66GHz Xeon 'Westmere' processors and a 24GB RAM. Our net delivered an accuracy of 92.68% (average) on the complete testing (which classifieds 4,538,439 effective points). The highest accuracy achieved on a single image was 94.54%, the lowest 88.85% (see **Table 4** for more details). It took in average **3750.55** seconds to completely segment an image. **Table 5** tabulates the confusion matrix on our four-class classification; **Table 6** presents the matrix in terms of percentage. **Figure 6** shows some of the segmentation output.

**Table 5.** Confusion matrix (in number of points).

| | Classification | | | | Total |
|---|---|---|---|---|---|
| | **Background** | **Optic disc** | **Fovea** | **Blood vessels** | |
| **True background** | 3,642,327 | 17,757 | 36,073 | 118,825 | 3,814,982 |
| **True optic disc** | 7,995 | 69,032 | 0 | 1,508 | 78,535 |
| **True fovea** | 6,829 | 0 | 59,295 | 853 | 66,977 |
| **True blood vessels** | 125,304 | 14,864 | 2,168 | 435,609 | 577,945 |
| | | | | | 4,538,439 |

**Table 6.** Confusion matrix (in percentage).

| | Classification | | | | Total |
|---|---|---|---|---|---|
| | **Background** | **Optic disc** | **Fovea** | **Blood vessels** | |
| **True background** | 95.47% | 0.47% | 0.95% | 3.11% | 100% |
| **True optic disc** | 10.18% | 87.90% | 0% | 1.92% | 100% |
| **True fovea** | 10.20% | 0% | 88.53% | 1.27% | 100% |
| **True blood vessels** | 21.68% | 2.57% | 0.38% | 75.37% | 100% |

**Table 7** tabulated the sensitivity and specificity in each classes. For a particular class, the sensitivity is calculated by *the number of pixels correctly classified as the class* divided by *the total number of pixels that belong to the class*. For specificity, it is calculated by *the number of pixels correctly classified as not the class* divided by *the total number of pixels that do not belong to the class*.

**Table 7.** Sensitivity and specificity in respective classes.

| Class | Sensitivity | Specificity |
|---|---|---|
| **True background** | 0.9547 | 0.8063 |
| **True optic disc** | 0.8790 | 0.9927 |
| **True fovea** | 0.8853 | 0.9914 |
| **True blood vessels** | 0.7537 | 0.9694 |

**Table 8** detailed the comparison of the performance of our algorithm against previous studies. We only included studies that reported sensitivity and specificity or overlap, evaluated on DRIVE database. For vasculature segmentation, we did not include all the literature in the table; we selected only some of the best results in each type of methods (according to the categories proposed by [1]).

It can be seen that in terms of sensitivity, our method performed reasonably well as compared to previous works, despite the fact that our algorithm was arranged to simultaneously segmented vasculature, optic disc and fovea, instead of only one or two of them. Our method has a lower specificity, which we think part of the reason, is because at the region of optic disc, some of the pixels that belong to optic disc were misidentified as blood vessels and vice versa.

**Table 8.** Comparison of performance measures on segmentation (not localization) against previous studies evaluated on DRIVE database.

| | Year | Blood vessels | | Fovea | | Optic disc | | |
|---|---|---|---|---|---|---|---|---|
| | | Sensitivity | Specificity | Sensitivity | Specificity | Sensitivity | Specificity | Overlap |
| **Our method** | **2017** | **0.7537** | **0.9694** | **0.8853** | **0.9914** | **0.8790** | **0.9927** | **0.6210** |
| 2nd observer **[30]** | | 0.7796 | 0.9717 | - | - | - | - | - |
| Zhu et al. **[2]** | 2016 | 0.7140 | 0.9868 | - | - | - | - | - |
| Wang et al. **[3]** | 2014 | 0.8104 | 0.9791 | - | - | - | - | - |
| Fraz et al. **[4]** | 2012 | 0.7406 | 0.9807 | - | - | - | - | - |
| Zhang et al. **[5]** | 2010 | 0.7120 | 0.9724 | - | - | - | - | - |
| Delibasis **[7]** | 2010 | 0.7288 | 0.9505 | - | - | - | - | - |
| Miri & Mahloojifar **[8]** | 2011 | 0.7352 | 0.9795 | - | - | - | - | - |
| Mendonca & Campilho **[9]** | 2006 | 0.7344 | 0.9764 | - | - | - | - | - |
| Vlachos & Dermatas **[10]** | 2010 | 0.747 | 0.955 | - | - | - | - | - |
| Martinez-Perez {MartinezPerez:2007bw} | 2007 | 0.7246 | 0.9655 | - | - | - | - | - |
| Li et al. **[12]** | 2007 | 0.780 | 0.978 | - | - | - | - | - |
| Espona et al. **[13]** | 2008 | 0.7436 | 0.9615 | - | - | - | - | - |
| Palomera-Perez et al. **[14]** | 2010 | 0.64 | 0.967 | - | - | - | - | - |
| Welfer et al. **[36]** | 2013 | - | - | - | - | 0.8354 | 0.9981 | 0.4254 |
| Stapor et al. **[37]** | 2004 | - | - | - | - | 0.7368 | 0.9920 | 0.3342 |
| Kande et al. **[38]** | 2008 | - | - | - | - | 0.6999 | 0.9888 | 0.2966 |
| Seo et al. **[39]** | 2004 | - | - | - | - | 0.5029 | 0.9983 | 0.3109 |
| Walter et al. **[40]** | 2002 | - | - | - | - | 0.4988 | 0.9981 | 0.2932 |
| Sopharak et al. **[38]** | 2008 | - | - | - | - | 0.2104 | 0.9993 | 0.1688 |

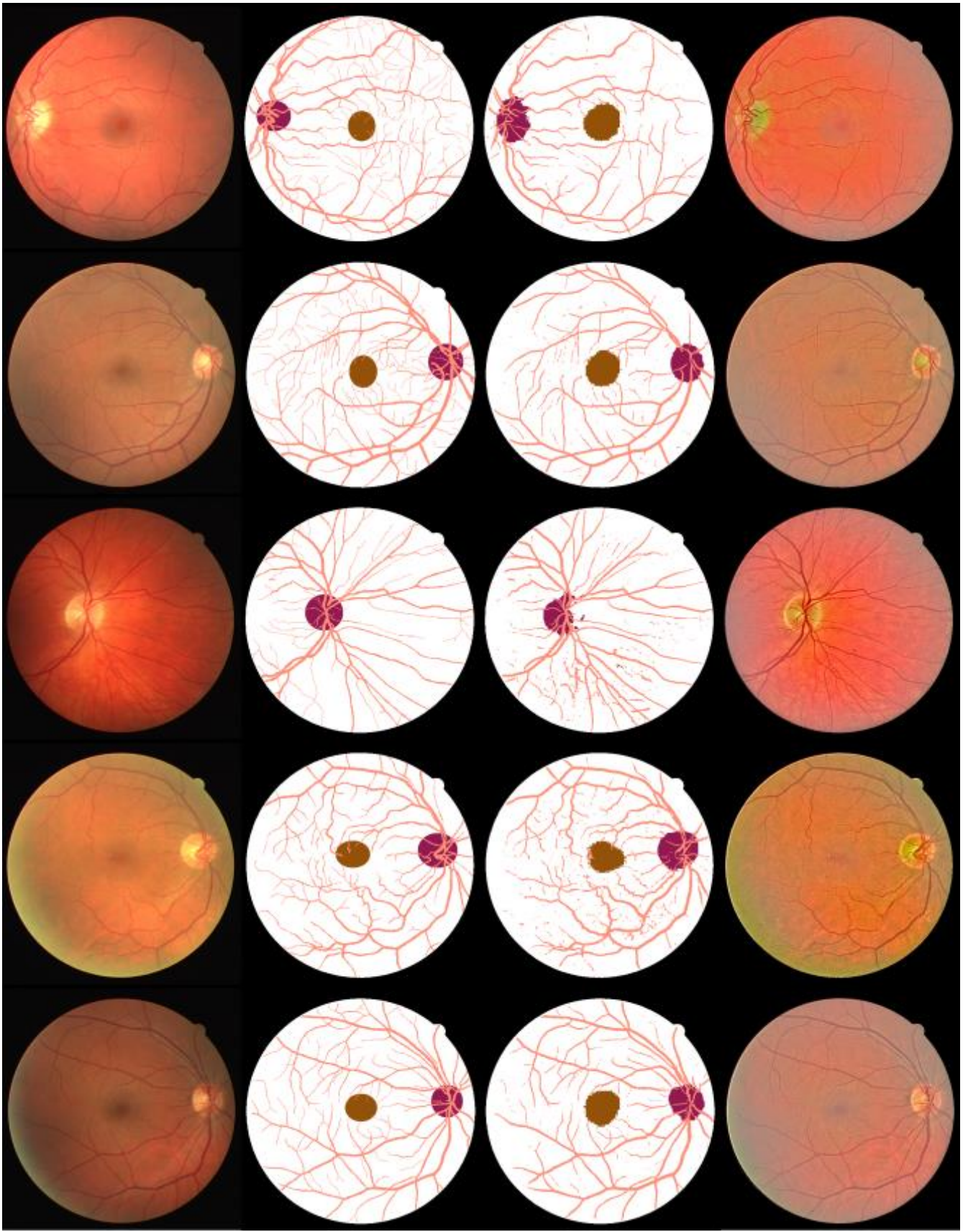

**Figure 6.** Segmentation results for image 11, 13, 15, 17 and 19. The images at the first, second, third, fourth columns are: original images, ground truths, segmentations and normalized images.

**5. Discussion**

Convolutional neural network has worked wonderfully on a number of image identification problems. It has been trained to discriminate Arabic numerals with almost perfect accuracy [41]. In terms of categorizing images (of 1000 classes), it even outperformed human observer [32].

But using a convolutional neural network to segment blood vessels, optic disc and fovea out of a fundus image is not the same as numeral recognition, or image categorization in at least three aspects. First, in our problem we perform segmentation by classifying the membership of **every single pixel**, whereas most of the other problems only identify the membership of an image, or a segment of image.

Second, the membership of a pixel can be ambiguous, especially along the boundary. Take for example, in **Figure 7a**, should the pixel at the middle (marked by a green dot) be classified as background or blood vessels? A boundary that is sharp and clear when viewed at its original scale does not necessarily imply a clear separation of classes at pixel level. When this happens, it is never easy to determine a pixel's membership with good confidence.

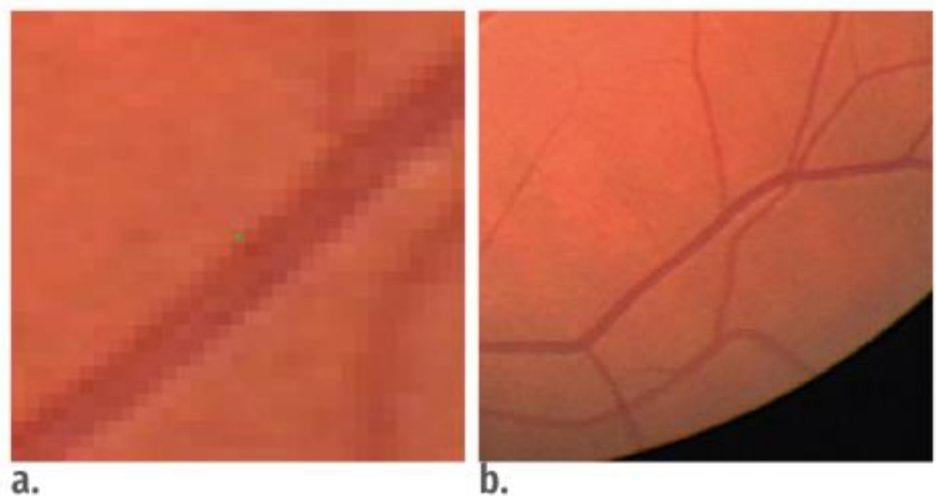

**Figure 7.** Ambiguity of membership at pixel level. a. The image is scaled up 10 times, and the green dot denotes the pixel of interest. b. The image is scaled up only 2 times at the same location.

Third, the membership of some pixels can be inferred **only** with reference to a larger neighbourhood around the pixel of interest. This is especially the case for optic disc and fovea, as illustrated in **Figure 2**. However, taking a large neighbourhood as input to neural network cannot be considered as good solution, for two reasons.

First, the memory requirement to train and test the net is excessive. Assume a frame of 101 x 101 is sufficient to discriminate all classes of pixels. A single such input extracted from a grey image will take 101 x 101 x 8 = 81,608 bytes (assume every pixel in an image is expressed by an unsigned integer, not double). If a computer is to hold the entire training set—which has 750,000 samples—in random access memory, it will need around 60 Gigabytes memory space, and this is not inclusive of another million samples required to test the net at the end of every training epoch!

Second, although a large neighbourhood is useful to determine the pixels of optic disc and fovea, majority of the pixels in the neighbourhood are unnecessary for the determination of the pixels of blood vessels and background. Wang et al. [3] has proposed a net that could accurately discriminate a pixel between classes of blood vessels and background using only a 25 x 25 input. Assume that size of input is what is only needed for the task, this will render nearly 94% of the 101 x 101 input useless to the separation of blood vessels and background. And since pixels of blood vessels and background make up almost 95% of the total effective pixels, it is rather inefficient to train and test a net where most of the information provided at input is redundant.

Instead, we have designed a net with an input of three channels—each distinctly scaled—to perform the segmentation. Channel 1 is put in to overcome the problem of ambiguity along boundary. It is inserted to push the net to learn extensively and exclusively on the pixels that are close to the point of interest, so that it can resolve the confusion arisen at the boundary.

Channel 3, on the other hand, is included to capture the macro pattern around the point of interest. The pixels of optic disc and fovea are relatively straightforward to determine if the macro pattern around the point of interest is available to the net. And the solution works, as illustrated in **Table 6**. For optic disc and fovea, 87.90% and 88.53% of their respective pixels were correctly identified, as compared to 75.37% for blood vessels (it is important to note that, however, blood vessels have much larger presence in fundus image).

Segmentation of blood vessels is important in the cardiovascular field and ophthalmic diseases. For example, morphological changes in retinal blood vessels are associated with cardiovascular disease [42] and hypertension [43]. Current tools to quantify these changes are cumbersome and time-consuming. The ability to automate the segmentation of blood vessels may help us to progress in the research on whether blood vessel changes may be a risk factor for cardiovascular disease. Venous dilatation and beading in diabetic retinopathy is again a useful tool for predicting progression [44] from severe non-proliferative diabetic retinopathy to proliferative retinopathy defined by the development of new blood vessels. This change in grade indicates the need to treat retinopathy. Therefore, accurate segmentation of blood vessel change may be used as a marker of progression if automated grading is utilized in diabetic retinopathy screening in future. Treatment response to panretinal photocoagulation [45] can also be monitored by blood vessel caliber changes if we can accurately define width of the blood vessels. In babies, the presence of plus disease is again defined by vessel width and tortuosity [46]. Like the iris, the retinal vascular tree is unique for each individual and may be used for biometric identification [47].

Segmentation of the optic disc is the first step towards automated screening and diagnosis of glaucoma [48]. The optic disc blood vessels and the peripapillary atrophy are challenges for accurate assessment of cup disc ratio even for a human observer. Defining the boundary of the optic disc is therefore crucial in automated segmentation.

The average diameter of the foveal avascular zone shows considerable inter-individual variation [49]. However, serial measurement of the foveal avascular zone is a predictive marker of macular ischemia [50]. Thus segmentation of the fovea will be very useful in annual screening of diabetic retinopathy.

Our algorithm so far took about 3751 seconds to complete segmentation on a single image. However that was achieved using only CPU. It is agreed that convolutional neural network runs faster on GPU, with probably 10 or 20 times quicker in processing speed [51]. In this case, our algorithm can complete the segmentation just under 5 minutes if it is run on a GPU, and much quicker if it is run on multiple GPUs.

Furthermore, unlike many other methods, our algorithm would not produce segmentation of optic disc or fovea when any of them is absent in a fundus image. It does not perform searching across the image by assuming optic disc or fovea is present in the image. We believe this is a desirable property that makes our algorithm more robust and versatile compared to many other techniques proposed in literature.

## 6. Conclusion

Most of the segmentation on fundus image can only segment/locate either vasculature, optic disc and fovea. We have developed an algorithm that can automatically and simultaneously segment all three of them. The algorithm consists of two parts: background normalization and 7-layer convolutional neural network. We have trained and tested our method on DRIVE database. It has achieved an average of 92.68% in terms of accuracy.

**Acknowledgement**

The research was supported by the National Institute for Health Research (NIHR) Biomedical Research Centre based at Moorfields Eye Hospital NHS Foundation Trust and UCL Institute of Ophthalmology.  The views expressed are those of the authors and not necessarily those of the NHS, the NIHR or the Department of Health.